\documentclass[letterpaper, 10 pt, conference]{ieeeconf}
\IEEEoverridecommandlockouts
\usepackage{graphicx}
\usepackage{amsmath}
\usepackage{amssymb}
\usepackage{censor}
\RestartCensoring

\begin{document}

\title{\LARGE \bf
Embodied Human Simulation for Quantitative Design and Analysis of Interactive Robotics}

\author{Chenhui Zuo$^{1}$, Jinhao Xu$^{1}$, Michael Qian Vergnolle$^{1}$, Yanan Sui$^{1}$%
\thanks{$^{1}$The authors are with the School of Aerospace Engineering, Tsinghua University, Beijing, China 100084. Correspondence to Yanan Sui. E-mail: {\tt\small ysui@tsinghua.edu.cn}}%
}

\newcommand{\model}{Digital Human Embodiment}

\maketitle
\thispagestyle{empty}
\pagestyle{empty}

\begin{abstract} 
Physical interactive robotics, ranging from wearable devices to collaborative humanoid robots, require close coordination between mechanical design and control. However, evaluating interactive dynamics is challenging due to complex human biomechanics and motor responses. Traditional experiments rely on indirect metrics without measuring human internal states, such as muscle forces or joint loads. To address this issue, we develop a scalable simulation-based framework for the quantitative analysis of physical human-robot interaction. At its core is a full-body musculoskeletal model serving as a predictive surrogate for the human dynamical system. Driven by a reinforcement learning controller, it generates adaptive, physiologically grounded motor behaviors. We employ a sequential training pipeline where the pre-trained human motion control policy acts as a consistent evaluator, making large-scale design space exploration computationally tractable. By simulating the coupled human-robot system, the framework provides access to internal biomechanical metrics, offering a systematic way to concurrently co-optimize a robot's structural parameters and control policy. We demonstrate its capability in optimizing human-exoskeleton interactions, showing improved joint alignment and reduced contact forces. This work establishes embodied human simulation as a scalable paradigm for interactive robotics design. 
\end{abstract}

\section{Introduction}

The human musculoskeletal system is an intricate biological machine, capable of remarkable dexterity and adaptation \cite{uchida2021biomechanics}. Its complexity, characterized by hundreds of muscles acting across a complex skeleton, is central to our ability to move and interact with the world. The design and control of interactive robotics, ranging from wearable exoskeletons to collaborative humanoid robots, show great potential to assist human motion and augment physical capabilities \cite{ghonasgi2024crucial, obayashi2025embodied}. However, the effectiveness of these technologies depends on a deep, quantitative understanding of how robotic assistance influences human biomechanics and motor control \cite{nanavati2023physically}.

Current evaluations rely heavily on human subject experiments \cite{zhang2017human,slade2022personalizing}. While these experiments are essential for validation, they are costly and time-consuming. Furthermore, they are fundamentally limited to measuring external and indirect metrics, such as metabolic cost or overall movement kinematics \cite{siviy2023opportunities}. These experiments cannot provide direct access to crucial internal biomechanical states, such as individual muscle forces or joint contact loads, which are essential for a comprehensive understanding of human-robot interaction \cite{lee2016essential, bobu2024aligning}. This limitation makes it exceedingly difficult to systematically evaluate how design and control affect the human body at a deeper biomechanical level.

\begin{figure}[h] 
\centering 
\includegraphics[width=\columnwidth]{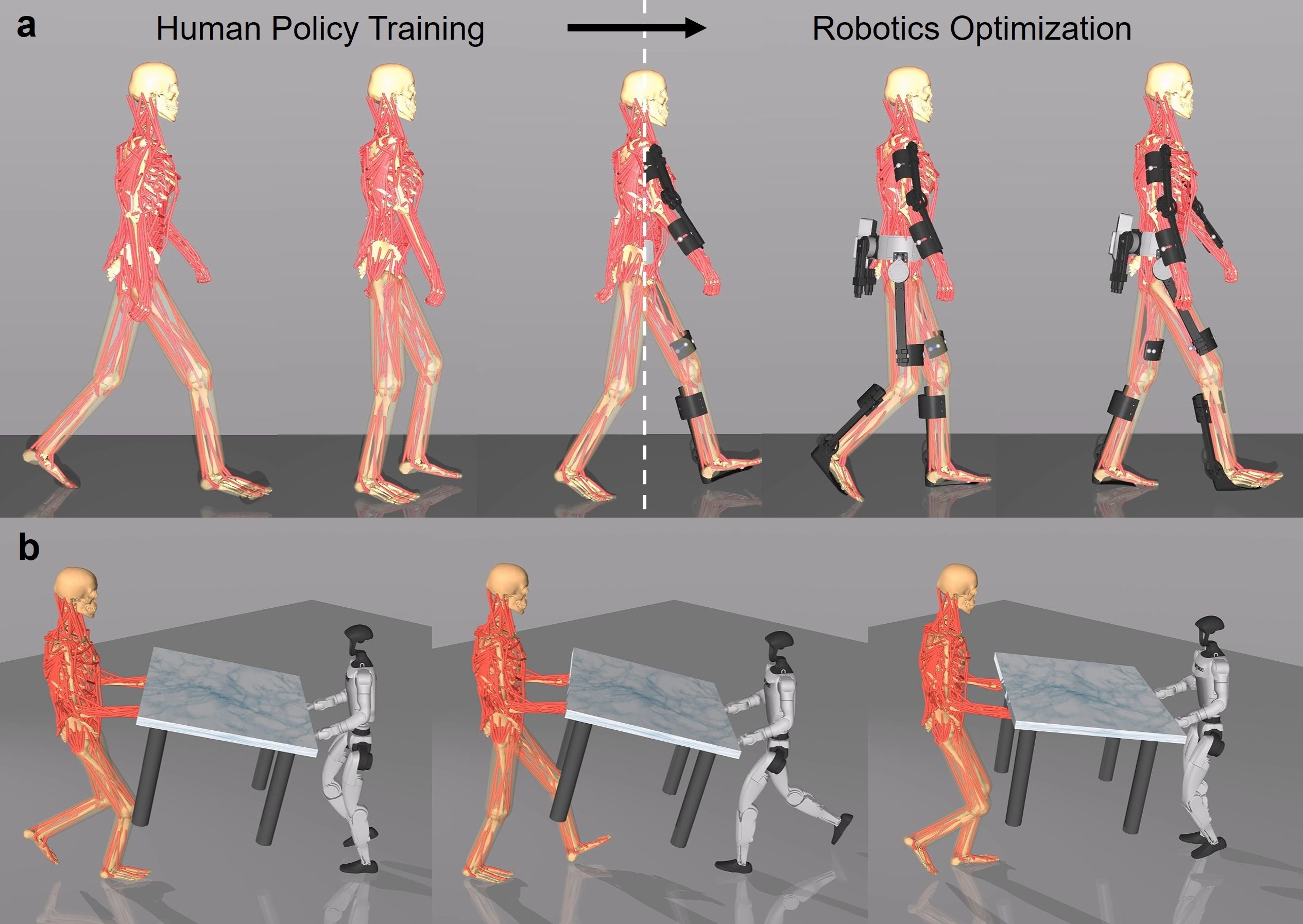} 
\caption{\textbf{Demonstrations of the \model~with interactive simulation framework.} (a) The co-optimization pipeline applied to a wearable exoskeleton, the human policy is trained and applied to robotics optimization. (b) The framework's scalability to diverse interactive robotics, illustrated by a daily collaborative task with a humanoid robot.} 
\label{fig:figure1} 
\end{figure}

Simulation-based methods provide an alternative that can overcome these barriers by enabling a rapid, safe, and cost-effective analysis \cite{fang2023human}. Recent studies have demonstrated the feasibility of developing control strategies entirely in simulation and transferring them to physical hardware, yielding significant performance improvements without the need for direct human testing \cite{luo2024experiment}. However, many existing simulation studies rely on simplified human models that fail to capture the complex, full-body neuromuscular dynamics essential to natural movement. A key limitation of these models is their inability to represent holistic coordination strategies. Without modeling full-body dynamics, it is difficult to create digital surrogates that provide physiologically grounded feedback during physical interactions. Consequently, the field lacks a unified computational platform that integrates an interactive, full-body human model for the safe and systematic analysis of physical human-robot interaction.

To address this critical gap, we present a simulation framework for the quantitative analysis of physical human-robot interaction. Our method's core is a comprehensive musculoskeletal model that includes the skeletons, joints, and muscle-tendon units of the entire human body. This model is actuated by a neural network controller that was trained using deep reinforcement learning \cite{zuo2024self, he2024dynsyn}. This digital human embodiment is not only capable of tracking specific motion trajectories, such as walking, but is also trained to exhibit robust, human-like responses to external physical disturbances. Within the simulation, we integrate models of interactive robotics and their physical coupling with the human body. By serving as a dynamic and interactive surrogate for a human user, our simulation provides access to a rich set of biomechanical and interaction metrics that are difficult or impossible to measure experimentally. This capability allows for a granular analysis of how a robot's design and control influence the user's internal biomechanics.

We demonstrate the framework through a case study involving a wearable exoskeleton \cite{williams2025openexo} for walking assistance. The study shows that detailed, quantitative feedback can be leveraged to guide the concurrent exploration of robot control policies and structural designs. Embodied human simulation establishes a scalable and human-centered paradigm for advancing interactive robotics. To support further research and development, we will open-source our framework, paving the way toward safer, more efficient, and more personalized assistive technologies.

\section{Related Work}

Human subject experiments, particularly human-in-the-loop (HIL) optimization, are the gold standard for tuning robot controllers using real-time physiological feedback \cite{zhang2017human, slade2022personalizing, ding2018human}. However, treating the human user as a black box presents significant limitations. While metabolic cost indicates overall effort, it obscures underlying biomechanical mechanisms, failing to reveal how an exoskeleton alters individual muscle recruitment strategies or joint loading patterns \cite{yan2015review}. Furthermore, acquiring ground-truth internal biomechanical labels (e.g., joint moments) for physical device testing is notoriously resource-intensive and device-specific \cite{scherpereel2025deep}. These time, cost, and safety constraints severely restrict the scope of evaluation to a small set of control parameters, leaving the vast design space of a robot's physical structure largely unexamined.

Human biomechanical simulation offers a powerful alternative to overcome these experimental barriers, providing a safe environment to probe internal metrics that are impossible to measure non-invasively \cite{delp2007opensim,molinaro2024task}. Recent research incorporates adaptive human models to optimize exoskeleton controllers \cite{gordon2022human,durandau2022neuromechanical,tan2025myoassist,leem2026exo}. A notable recent breakthrough demonstrated experiment-free learning, where controllers trained entirely in simulation were successfully transferred to physical hardware without any HIL tuning \cite{luo2024experiment}.

Despite this progress, a critical limitation persists: existing frameworks typically optimize a control policy for a predefined, fixed hardware design. The concurrent exploration of a robot's structural parameters alongside its control system remains a largely unaddressed challenge. Furthermore, previous work often employs simplified human models, limiting the replication of complex, full-body neuromuscular dynamics \cite{rajagopal2016full, lee2019scalable, erickson2020assistive, upaddhye2025reinforcement}. Our work builds on this progress by presenting a unified computational framework that employs a comprehensive, full-body musculoskeletal model. By doing so, we enable the systematic, concurrent exploration of the both structural and control parameters, which reveals the intricate relationships between a robot's physical form and its biomechanical impact on the human user.

\section{Modeling the Coupled Human-Robot System}

\subsection{Full Body Human Musculoskeletal Model}

\begin{figure}[h]
    \centering
    \includegraphics[width=1\columnwidth]{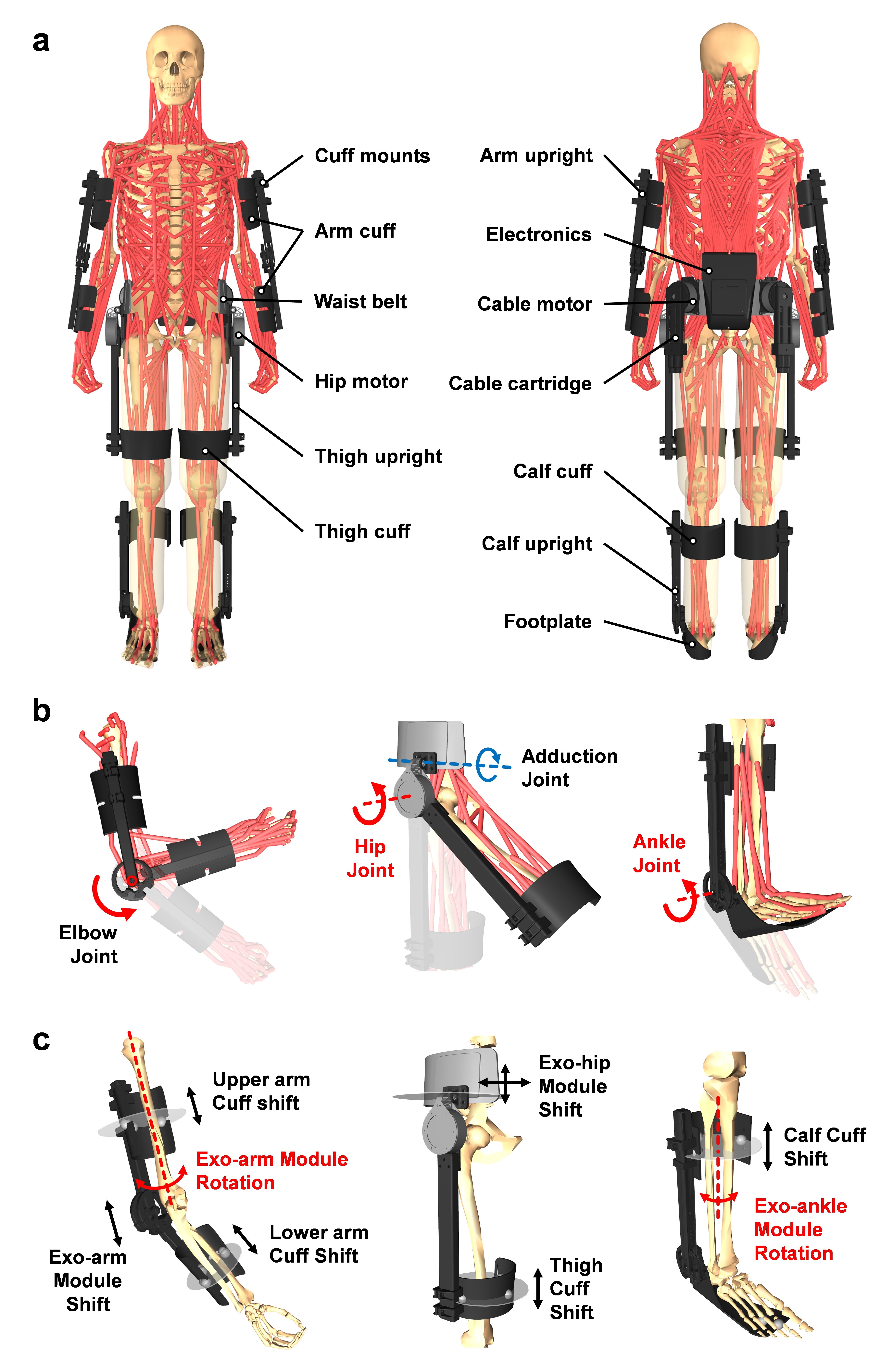}
    \caption{\textbf{The human-exoskeleton coupled simulation model.} (a) The full system, showing the MS-Human-700 model wearing the OpenExo exoskeleton. (b) A detailed view of the human-robot physical interface, where the adduction joint is passive to allow adaptation to natural leg abduction movement. (c) Illustration of the adjustable structural parameters in the interactive system. Local cuff adjustments correspond to vertical shifts at binding sites, while global assembly module adjustments modify the axis relative to attached human body segment through rotations and translations. The interactions were modeled with compliant elastic tendons elements (gray spheres) that connect the exoskeleton shell to the underlying human segment and transmit assistive forces.}
    \label{fig:exo_interaction}
\end{figure}

Our framework is built upon the MS-Human-700, an open-source full-body musculoskeletal model comprising 90 rigid body segments, 206 joints, and 700 individual muscle-tendon units \cite{zuo2024self}, which is shown in Figure \ref{fig:exo_interaction}a with an coupled exoskeleton. This model is implemented within the MuJoCo physics engine \cite{todorov2012mujoco}, which simulates its complex dynamics. To capture the system's realism, the simulation incorporates established biomechanical principles. The overall dynamics of the skeleton are governed by the Euler-Lagrangian equation: \begin{equation} 
M(q)\Ddot{q} + c(q,\Dot{q}) = J_{m}^T f_m(act) + J_c^T f_c + \tau_{ext} 
\end{equation} 
Here, $\Ddot{q}$ represents joint accelerations, driven by internal muscle forces ($f_m$), external interaction torques ($\tau_{ext}$), and gravity/Coriolis effects $c(q,\Dot{q})$. $M(q)$ denotes the mass matrix, while $J_m$ and $J_c$ are the respective Jacobians. The biomechanically validated Hill-type muscle model \cite{millard2013flexing} is employed for the 700 actuators. The muscle force ($f_m$) is a nonlinear function of activation ($act$), fiber length ($l_m$), and contraction velocity ($v_m$): 
\begin{equation} 
f_m(act)=f_{max}\cdot [F_{l}(l_m)\cdot F_v(v_m)\cdot act + F_p(l_m)] 
\end{equation}
where $F_l$, $F_v$, and $F_p$ capture the force-length, force-velocity, and passive elastic relationships. Furthermore, the model includes first-order activation dynamics to represent the physiological delay between a neural control signal and the resulting muscle activation, using state-dependent time constants.

While controlling this complex system is challenging, its comprehensive structure is essential for capturing physiologically grounded dynamics. The anatomical pathways and physiological properties of the muscle-tendon units provide natural constraints on movement and force generation. This approach captures the full-body coordination inherent in human motion, such as the role of the torso and arms in stabilizing walking. These features are critical for creating a predictive digital surrogate that responds to robotic interaction more reasonably than simplified torque-driven models.

\subsection{Modeling the Interactive Exoskeleton}

To quantitatively evaluate physical interactions, we construct a fully parameterized coupled human-robot model in the simulation. While our framework is highly generalizable and supports simulating complex interactions with collaborative humanoid robots (as shown in Figure \ref{fig:figure1}), we select a wearable exoskeleton based on the OpenExo platform \cite{williams2025openexo} as our primary case study. This choice highlights the ability of our method for analyzing a wide range of full-body assistive applications.

The physical structure of the robot is modeled as a rigid-body system within the MuJoCo physics engine. As shown in Figure \ref{fig:exo_interaction}, the exoskeleton has actuated joints at the hips, ankles, and elbows. Based on the assembly weights reported by OpenExo, we assigned density properties to certain parts. This enabled a more reasonable dynamic description of the interactive device in simulation and provides a basis for further customization. 

The critical element of the simulation is the physical interface that connects the exoskeleton to the human model. We model this interface using compliant connectors that simulate the viscoelastic behavior of soft straps or cuffs on a real device \cite{serrancoli2019subject, massardi2022characterization}. In our implementation, these connectors are defined as spatial tendons acting as parallel spring-damper systems in MuJoCo. They link attachment points on the exoskeleton frame to corresponding points on the user's body segments. Unlike rigid weld constraints that often cause unrealistic interaction force spikes during kinematic mismatches, these parameterized compliant tendons effectively transmit assistive torques while permitting natural micro-misalignments. This provides a much more physiologically accurate estimation of peak contact forces.

A significant advantage of our simulation-based framework is the ability to programmatically adjust the exoskeleton's structural parameters. We achieve this by parameterizing key structural properties of the robot model, such as the positions of attachment sites or the relative orientation of entire modules. Figure \ref{fig:exo_interaction}c illustrates several of these adjustable structural parameters. These adjustments include local changes, such as shifting the position of the thigh and calf cuffs along the robot's links to alter where forces are applied. They also include global changes, such as translating the entire hip module or rotating the ankle and arm modules. These global adjustments modify the alignment between the robot's joints and the user's biological joints. By including these parameters in the optimization space alongside the controller, our framework can explore synergistic relationships between the robot's physical form and its control strategy.

\begin{figure*}[h]
    \centering
    \includegraphics[width=1.7\columnwidth]{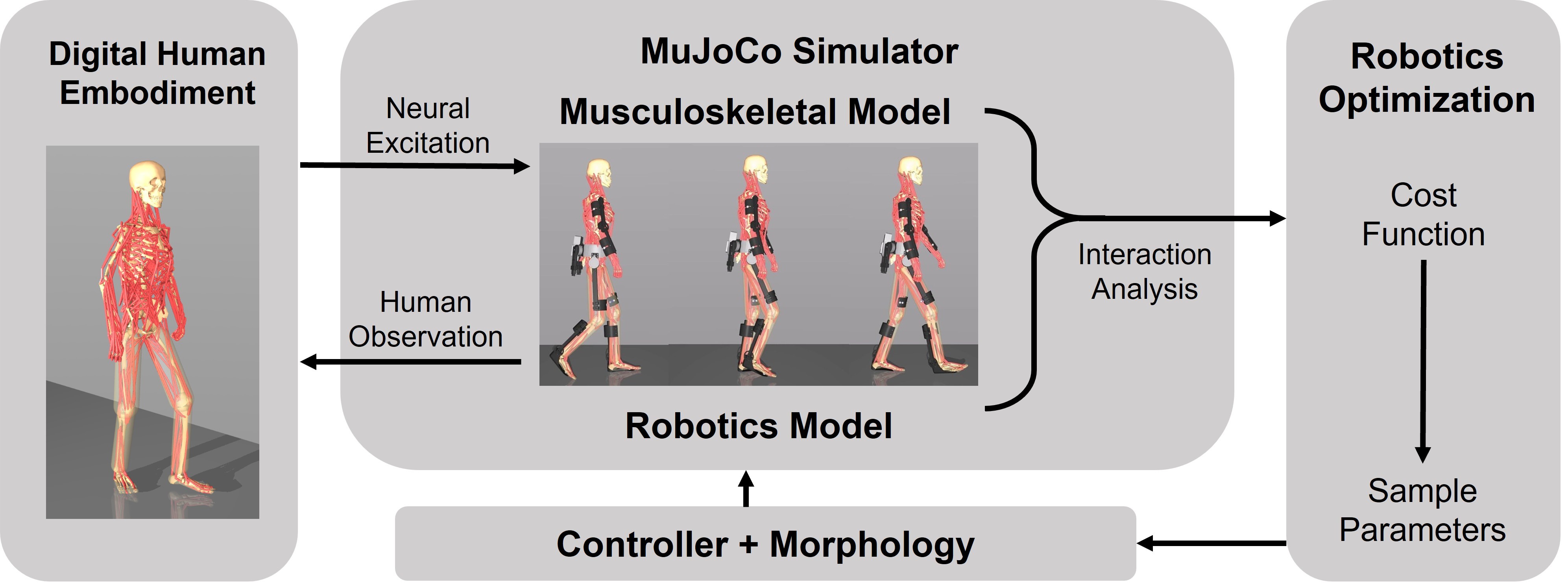}
    \caption{\textbf{Flowchart of the co-optimization loop.} The optimization algorithm proposes a new set of exoskeleton parameters (control and structure). The human-robot coupled simulation is executed for several gait cycles with these parameters. The resulting motion and physiological data are used to evaluate the cost function. This cost is returned to the optimizer, which updates its internal model and proposes the next set of parameters to evaluate, iterating until convergence.}
    \label{fig:optimization_flowchart}
\end{figure*}

\section{Learning Adaptive Motor Control for Interaction}\label{Sec:training}

\subsection{Training an Interactive Human Motor Policy}

We employ deep reinforcement learning (RL) to train the human musculoskeletal model that acts as a realistic, interactive surrogate for a specific human user performing a given motion task. This involves developing a task-specific control policy for the full-body musculoskeletal model, enabling it to robustly execute target motions while responding adaptively to external physical interactions. For demonstration in the case study, we focus on walking. Although walking primarily involves the lower limbs, simulating the full-body dynamics is crucial. Upper-body movements, including arm swing and torso stabilization, are integral to natural gait balance and are essential for capturing holistic, compensatory biomechanical responses to localized robotic assistance. Furthermore, this comprehensive full-body control directly supports the framework's scalability to upper-limb-dominant interactions, such as the collaborative manipulation task with a humanoid robot illustrated in Figure \ref{fig:figure1}.

To address MS-Human-700 model's high dimensionality and redundant biomechanics, we employ the DynSyn \cite{he2024dynsyn} algorithm based on Soft Actor-Critic \cite{haarnoja2018soft}, a method designed for efficient learning in overactuated systems. DynSyn operates on the principle of muscle synergies, a neuroscientific theory suggesting the central nervous system simplifies motor control by coordinating groups of muscles to act as functional units \cite{grillner1985neurobiological, ting2007neuromechanics}. The algorithm generates synergistic representations of the actuators from the model's structure and then uses these representations to guide the learning process. Specifically, the policy and value networks are multi-layer perceptrons (MLPs) with three hidden layers (1024 units each). This bio-inspired approach improves sample efficiency and produces coordinated muscle activation patterns that more closely resemble natural human movement, which is critical for simulating realistic interaction feedback.

We formulate the RL problem as a Markov Decision Process. The action space $\mathcal{A}$ corresponds to neural excitations driving the 700 muscle-tendon units. The high-dimensional observation space $\mathcal{S}$ (3601 dimensions) includes joint kinematics, muscle states (activation, length, force), key body segment spatial positions, and reference kinematic targets \cite{peng2018deepmimic}. To guide the learning, the reward function $r(s,a)$ is designed as a weighted sum of multiple components to encourage kinematically accurate and biomechanically plausible motion: 
\begin{align} 
r_{joint} &= - \sum_{i} (q_i - q_{ref,i})^2 \nonumber \\ 
r_{position} &= - \sum_{j \in K} | p_j - p_{ref,j} |^2 \nonumber 
\\ r_{energy} &= - \sum_{i=1}^{N_a} u_i^2 \nonumber 
\end{align} 
Here, $q_i$ and $p_j$ are simulated joint angles and key body positions, while the subscript ref denotes target values from motion capture data. $r_{joint}$ and $r_{position}$ enforce kinematic fidelity. Biomechanical efficiency is driven by $r_{energy}$, which penalizes squared muscle excitations ($u_i$). An additional survival bonus $r_{healthy}$ is provided for remaining upright.

To enhance the robustness and generalizability of the learned policy, we incorporated several key design choices into the training process \cite{luo2023perpetual, liao2025beyondmimic}. First, to prevent overfitting to a single idealized motion, training is performed on the set of ten trajectories. At the beginning of each episode, one of these trajectories is chosen at random. Second, we use random initialization: each episode starts at a random time point within the first trajectory cycle, with small noise added to the initial state. This ensures the agent learns to perform and recover the motion from any phase, not just from a static start. Finally, we employ an early termination condition. If the agent's joint angles deviate significantly from the reference motion, the episode is terminated. This strategy focuses the training on successful attempts and accelerates learning by avoiding time spent in unrecoverable states.

A key feature of our \model~is its ability to interact dynamically with its environment. Simply tracking a trajectory is insufficient for a pHRI surrogate, as it must respond to and recover from unexpected forces and torques. To build this robustness, we incorporate domain randomization into the training \cite{mehta2020active}. Throughout the learning process, we apply random, transient external forces to the model’s torso and limbs. This forces the policy to learn active recovery strategies that stabilize the body, rather than merely memorizing kinematic sequences.

\subsection{Parameterized Controller for Robot Assistance}

The robot's actuation is governed by a parameterized controller that determines the torque outputs at each joint. We employ a continuous, trajectory-tracking controller designed to assist the user in following a reference walking motion. The controller’s primary goal is to support the user’s natural movement patterns, reducing effort without constraining their preferred gait. We implement a decoupled Proportional-Derivative (PD) controller. While a standard PD controller computes torque based on combined position and velocity errors, our design separates feedforward (reference) and feedback (measured) terms. This decoupling enriches the optimization space, enabling the discovery of more subtle and effective assistance strategies. The assistive torque $\tau$ for each joint is calculated using the following formula:
\begin{equation}
\tau = K_{pr} q_{ref} - K_{py} q + K_{dr} \dot{q}_{ref} - K_{dy} \dot{q}
\end{equation}

Here, $\tau$ denotes the assistive torque generated by the exoskeleton motors. $q_{ref}$ and $\dot{q}_{ref}$ represent the target joint angle and angular velocity from reference motion capture data, while $q$ and $\dot{q}$ denote the current measured joint angle and velocity of the human model. The gains $K_{pr}$ and $K_{dr}$ correspond to the feedforward components for position and velocity, and $K_{py}$ and $K_{dy}$ are the feedback gains for the measured position and velocity, respectively.

This entire system operates in a unified, closed loop. At each step of the simulation, the neural network policy determines the appropriate muscle activations, while the exoskeleton's controller calculates the necessary motor torques based on its parameters and the human's current state. These forces, from both muscles and motors, are applied simultaneously, and the resulting motion is computed by the physics engine. The forces generated by the exoskeleton are transmitted to the human through the compliant tendon interface, causing the \model~to dynamically adapt its muscle recruitment in response. This creates a comprehensive testbed where the coupled human-robot interaction provides rich biomechanical data to guide the optimization process.

\section{Optimizing Robot Design and Control on Human Simulation Feedback}\label{Sec:optimization}

With the fully parameterized interactive human-robot model established, we now describe the optimization pipeline that leverages this simulation to automatically discover device designs. As illustrated in Figure \ref{fig:optimization_flowchart}, the pipeline automates the search for an optimal device design by treating the complex human-robot simulation as a cost function to be optimized.

Crucially, to make the vast design space exploration computationally tractable, we employ a sequential training-optimization pipeline. The robust human motor policy trained in Section \ref{Sec:training} is frozen and acts as a consistent physiological evaluator. The optimization process begins with an algorithm proposing a set of parameters that define the exoskeleton's controller gains and its structural parameters. These parameters are applied to the robot model, and the simulation runs for multiple gait cycles with the fixed agent generating neural excitations. By not forcing the human policy to co-adapt during optimization, we inherently search for robot designs that best accommodate the user's natural, pre-existing motor habits.

The resulting biomechanical data is evaluated by a multi-objective, human-centric cost function $C$ that quantifies the design performance. The overall objective is formulated as: 
\begin{equation} 
C = w_1 C_{kinematic} + w_2 C_{effort} + w_3 C_{interaction} 
\end{equation} 
The primary objective $C_{kinematic}$ penalizes deviations from the reference human motion, ensuring the device assists without disrupting the user's natural gait. The secondary objective $C_{effort}$ estimates metabolic cost by integrating the sum of squared muscle forces over the gait cycle, promoting physical energy reduction. The tertiary objective $C_{interaction}$ minimizes the peak contact forces at the physical interface (tendons) to serve as a proxy for user comfort. 

To efficiently navigate the high-dimensional, derivative-free design space, we employ the Covariance Matrix Adaptation Evolution Strategy (CMA-ES) \cite{hansen2006cma}. The optimization space is defined by a bounded 21-dimensional parameter vector to ensure physically plausible designs. This comprises 12 dimensions for the controller gains of the six actuated joints (bilateral hip, ankle, and elbow, with four tunable gains each from the decoupled PD controller) and nine dimensions governing the exoskeleton’s structural parameters (cuff positions and module transformations). To maintain balance, all structural adjustments are applied symmetrically to both sides of the body. CMA-ES iteratively evaluates populations of these parameter sets in the coupled simulation to minimize the cost function and converge on an optimal synergistic design.

Regarding computational efficiency, our framework leverages the speed of the MuJoCo engine. Training the robust human motor policy takes approximately 8 hours on a single NVIDIA RTX 4070 Ti GPU and 128 parallel CPU simulation environments. The subsequent CMA-ES co-optimization evaluates 300 of structural and control candidates. Each evaluation rollout takes only 10 seconds, leading to a total optimization convergence time less than one hour.

\section{Experiments and Results}

\subsection{Validation of the Digital Human Embodiment}

A prerequisite for meaningful human-robot simulation is a human model that exhibits realistic behavior. We therefore validated our trained agent by evaluating its unassisted walking patterns against biomechanical data and testing its stability against external perturbations.

To ensure applicability, the agent learns to walk by imitating human motion capture data. For our demonstration, we selected a representative medium-speed walking trajectory (closest to the mean velocity and stride length of the subject's trials) from the public AMASS database \cite{mahmood2019amass}. This serves as the reference motion for all subsequent evaluations, as it represents a typical assistive walking scenario.

\begin{figure}[h]
    \centering
    \includegraphics[width=1\columnwidth]{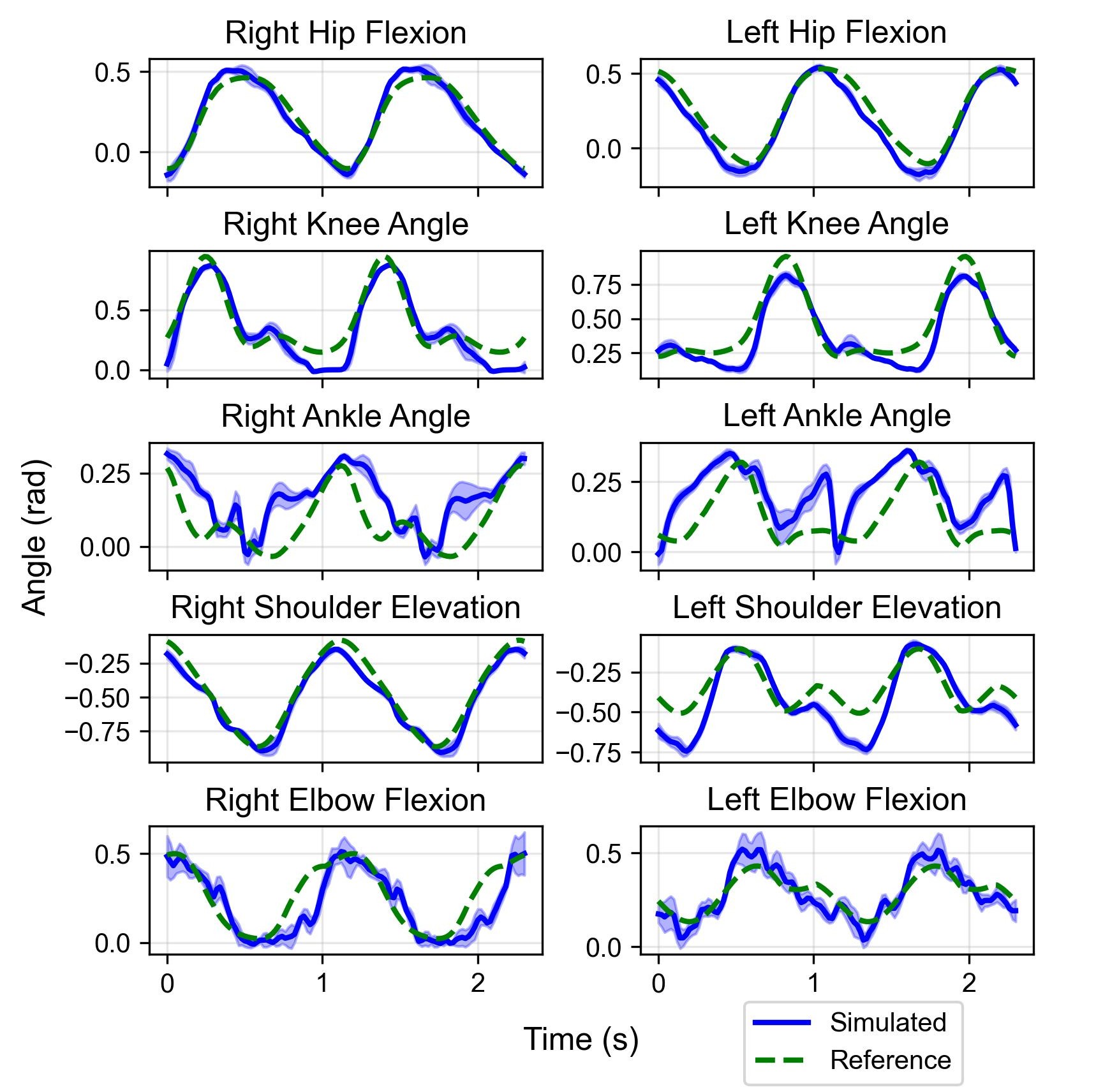}
    \caption{\textbf{Validation of the digital human's walking kinematics.} The simulated joint angles (blue solid lines) are compared with the reference motion capture data (green dashed lines) for major joints. The shaded areas represent the standard deviation across 10 simulation trials with different initialization time steps, indicating high consistency.}
    \label{fig:joint_angle_walking}
\end{figure}

We performed a quantitative comparison of the agent's learned walking gait against the reference motion capture trajectory. Figure \ref{fig:joint_angle_walking} displays this comparison for key joints across both the lower and upper body, including hip flexion, knee angle, ankle angle, shoulder elevation, and elbow flexion for both the left and right sides. The plots demonstrate a high degree of fidelity, as the simulated joint angle trajectories closely track the reference data throughout the gait cycles. The overall shape, timing, and amplitude of the motion are well-replicated for all major joints. The agent's joint kinematics matched the human walking data with an average root-mean-square (RMS) error below 0.05 rad. Minor differences between the simulated and reference walking patterns arise mainly because the musculoskeletal model is generalized and not tailored to the specific subject from whom the motion data was sourced. And the simulation's physical parameters may differ from real-world motion capture conditions. Despite these small mismatches, our learning-based controller still provides realistic full-body gaits for studying neuromuscular responses to interaction, establishing the agent as a credible base for subsequent studies.

\begin{figure}[h]
    \centering
    \includegraphics[width=1\columnwidth]{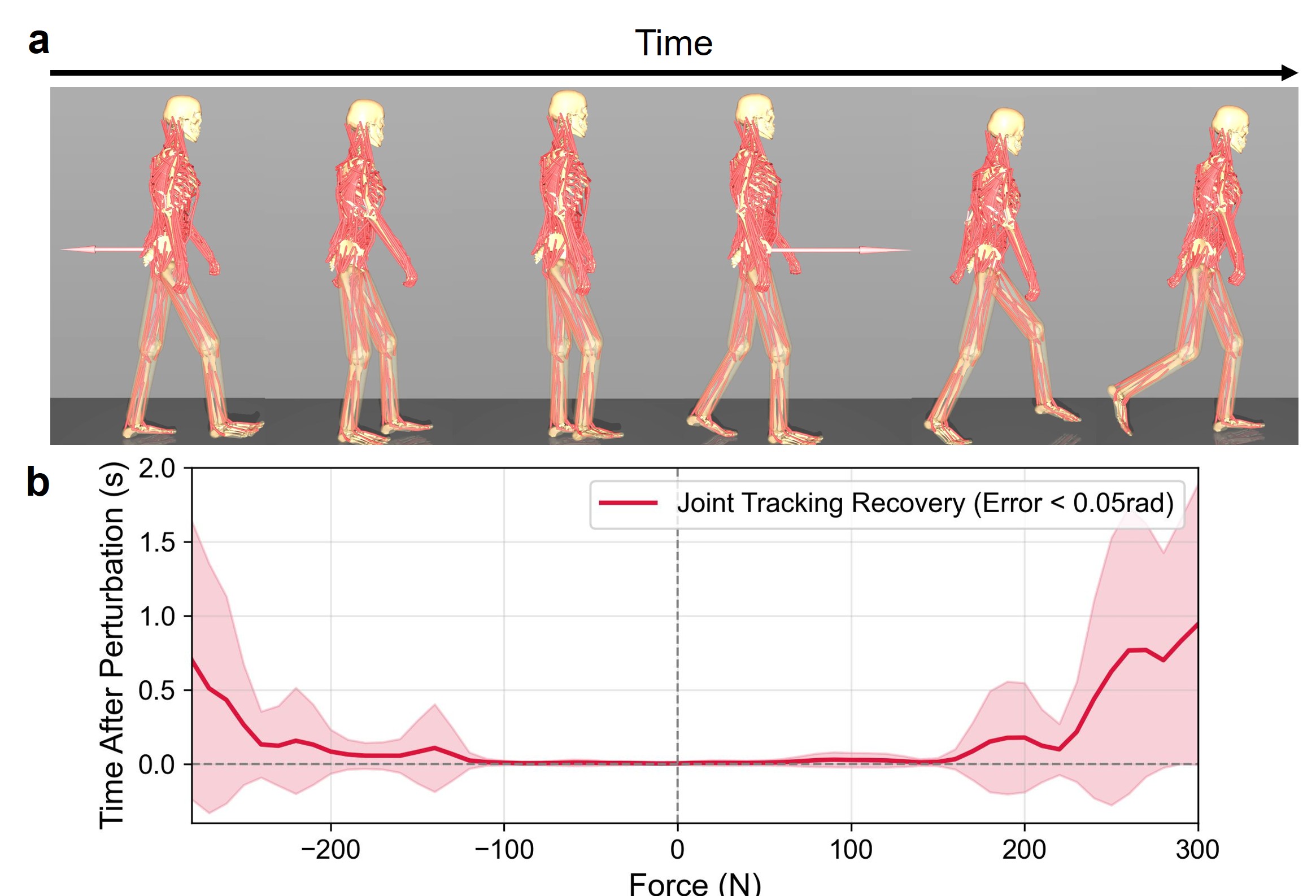}
    \caption{\textbf{Robustness evaluation of the learned walking policy against external perturbations.} (a) The agent's dynamic recovery sequence after an impulsive force is applied to the pelvis, where the pink arrow visualizes the applied disturbance force. (b) Recovery time across a range of forward (positive) and backward (negative) forces. The red line shows the time required to recover the original kinematic trajectory, demonstrating a fast return to the intended motion. The shaded area represents the standard deviation over 10 randomly initialized trials.}
    \label{fig:robustness_test}
\end{figure}

Next, we evaluated the robustness of the learned walking policy by subjecting it to unexpected impulsive forces applied to the pelvis, simulating external disturbances. To ensure a representative evaluation, we applied these disturbance forces for a duration of 0.2 seconds in both forward and backward directions. The agent demonstrated effective recovery strategies. Figure \ref{fig:robustness_test}a provides a visual example, showing the agent successfully restabilizing its gait and continuing to walk after significant forward and backward pushes in the pelvis. Figure \ref{fig:robustness_test}b quantifies this stability across a range of forces. The agent showed rapid kinematic recovery, returning to its reference trajectory with a joint tracking error of less than 0.05 rad in under 0.5 seconds for forces up to 200 N. 

These results confirm that the agent can robustly execute the target walking motion while actively withstanding external perturbations. This provides closed-loop biomechanical feedback in response to forces from a simulated interactive robotic device. This dynamic capability qualifies the agent as a scalable and capable surrogate for a human user, enabling the detailed, quantitative analysis of the coupled human-robot system.

\subsection{Optimization of Exoskeleton Design and Controller}

\begin{figure}[h]
    \centering
    \includegraphics[width=\columnwidth]{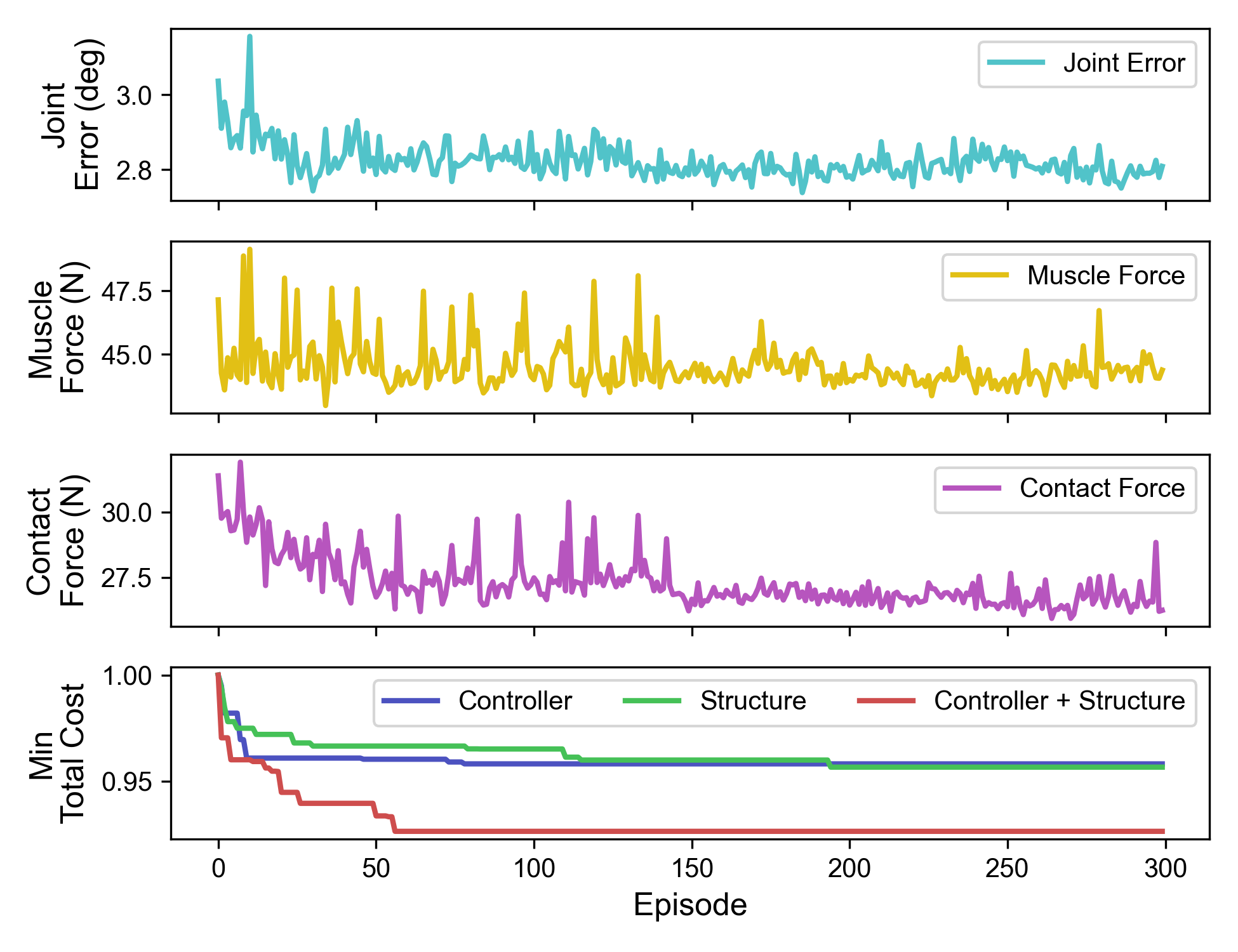}
    \caption{\textbf{Performance comparison of optimization strategies.} The top three panels illustrate the average joint error, muscle force, and contact force during the co-optimization of controller and structure parameters across episodes, showcasing the improvements. The bottom panel compares the normalized minimum total cost achieved so far for three strategies: controller-only (blue), structure-only (green), and co-optimization (red). The co-optimization approach demonstrates significantly faster convergence and a lower final cost, highlighting the benefits of concurrently optimizing control and structure.}
    \label{fig:optimization_results}
\end{figure}

Having established the credibility of our \model~as a realistic surrogate, we now demonstrate its utility in a practical design optimization task. In this section, we apply our framework to discover optimal exoskeleton parameters and reveal the synergistic benefits of co-optimizing the robot's control and physical form. We compared the performance of our co-optimization of control and structure against two strategies: (1) Control-Only Optimization, where structure was fixed and only control parameters were optimized; and (2) Structure-Only Optimization, where control was fixed and only structural parameters were optimized. The optimization started with a standard set of structural and control parameters. The results, summarized in Figure \ref{fig:optimization_results}, show that the co-optimization strategy yielded a substantially greater reduction in the total cost than any of the baseline conditions. While optimizing a single domain provided moderate benefits, the concurrent optimization of both structural and control parameters unlocked synergistic improvements.

\begin{figure}[h]
    \centering
    \includegraphics[width=\columnwidth]{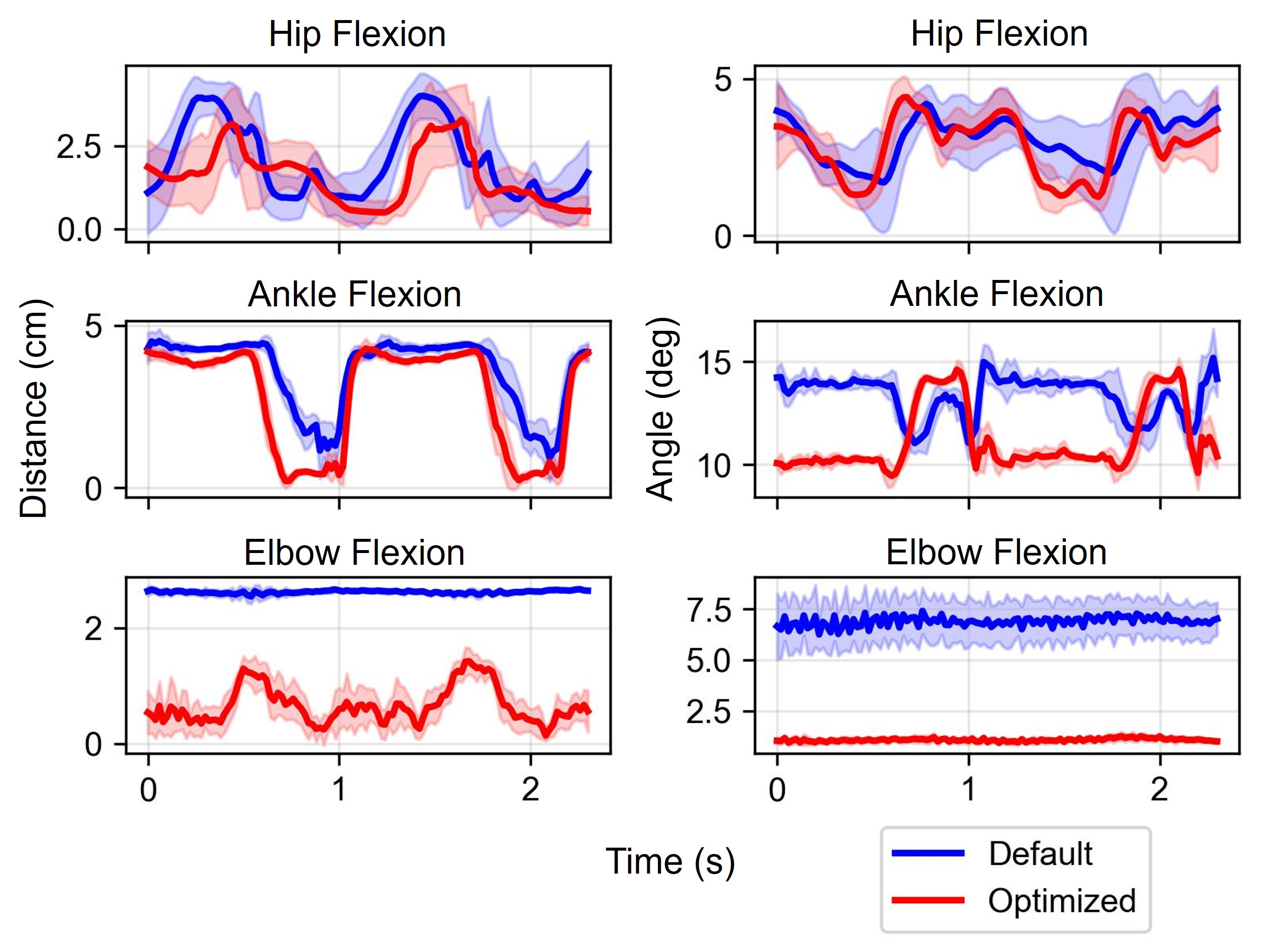}
    \caption{\textbf{Enhanced human-robot joint axis alignment after co-optimization.} The left column shows the shortest distance between the corresponding human and exoskeleton joint axes, while the right column shows the angle between them. Shaded areas represent the standard deviation across 10 simulation trials. The optimized parameters significantly reduce both distance and angular deviation, enabling more effective and comfortable force transmission.}
    \label{fig:joint_axis_alignment}
\end{figure}

To uncover the mechanisms behind this improvement, we analyzed how the optimized parameters altered the physical interaction between the human and the robot. A key factor in effective assistance is the alignment between the exoskeleton's joints and the user's biological joints. Misalignment can lead to inefficient force transmission and user discomfort. Our framework allows us to directly quantify this alignment by measuring the geometric relationship between the joint axes in the simulation. Figure \ref{fig:joint_axis_alignment} shows a comparison of joint axis alignment for the hip, ankle, and elbow joints during a walking cycle. The plots display two key metrics: the shortest distance between the corresponding human and exoskeleton joint axes (left column) and the angle between them (right column). The results clearly show that the co-optimized design achieves significantly better alignment than the default configuration (visually aligned by human). For all three joints, the optimized structure reduces both the distance and the angular deviation throughout the gait cycle.

This improved alignment is a direct outcome of including structural parameters in the optimization. The algorithm discovered adjustments to the cuff positions and module orientations that bring the robot's structure into closer harmony with the user's anatomy. As a result, the exoskeleton can apply assistive torques more precisely around the intended biological axes. This not only enhances the effectiveness of the assistance, leading to reduced metabolic cost, but also significantly decreases peak interaction forces at the tendon interfaces that could cause discomfort, directly addressing the $C_{interaction}$ objective in our cost function. This analysis highlights the framework's ability to reveal non-intuitive design choices that connect a robot's physical form to its biomechanical impact.

\section{Conclusion and Discussion}

This paper presents an embodied human simulation framework for quantitatively analyzing human-robot interaction. Our framework uses a comprehensive, full-body musculoskeletal model actuated by a learned neural network controller to provide a predictive surrogate that reveals critical biomechanical states that would otherwise be inaccessible, such as internal muscle activations and joint contact forces. Through a co-optimization case study on assistive walking, we demonstrated how this computational framework can guide concurrent, automated exploration of both robot structure and control. This establishes a new, highly scalable paradigm that moves beyond treating the human body as an unknown entity, enabling the systematic evaluation of complex design spaces that traditional human-in-the-loop experiments cannot easily access.

This simulation-driven approach has a clear pathway to real-world deployment. Analytical insights gained from simulation can directly inform hardware prototypes, as optimized design parameters and control policies can be directly transferred to physical devices. Embodied human simulation has the potential to reduce the time it takes to transition from conceptual models to high-performance interactive systems.

We should also discuss the scope and limitations of this work. While the walking example in this study was trained using data from one subject performing one task, our proposed framework is inherently generalizable. Our method can be trained on motion capture data from diverse individual performing different movement tasks. This process can be used to generate new digital human embodiments, enabling applications ranging from industrial collaboration to rehabilitation robotics. Nonetheless, the inherent sim-to-real gap remains a key challenge. Currently, our validation is primarily kinematic due to the lack of open-source biomechanical datasets containing synchronized full-body kinematics, EMG, or ground reaction forces for human-robot interactive tasks. Complexities such as soft tissue dynamics or user fatigue are not yet fully captured, meaning that comprehensive human subject experiments remain essential for final validation and rigorous physiological verification.

Future directions include expanding the library of digital human models to capture inter-subject variability, incorporating richer physiological models, and refining transferability to physical systems. Recent advances in deep domain adaptation show promise in translating simulated biomechanical sensor data to real-world wearable sensors \cite{scherpereel2025deep}, which could seamlessly connect our structural discoveries to physical deployment. These efforts aim to establish embodied human simulation as a cornerstone methodology for safe, scalable, and personalized human-robot interaction.

\bibliographystyle{IEEEtran}
\bibliography{sample-base}

\end{document}